\documentclass[sigconf]{acmart}

\AtBeginDocument{%
  }

\setcopyright{acmlicensed}
\copyrightyear{2026}
\acmYear{2026}
\setcopyright{cc}
\setcctype{by}
\acmConference[HAI '26]{Proceedings of the 14th International Conference on Human-Agent Interaction}{November 16--19, 2026}{Osaka, Japan}
\acmBooktitle{Proceedings of the 14th International Conference on Human-Agent Interaction (HAI '26), November 16--19, 2026, Osaka, Japan}
\acmDOI{10.1145/3841580.3841619}
\acmISBN{979-8-4007-2575-3/2026/11}

\begin{document}

\title{JuryFlow: Disagreement-Guided Human-in-the-Loop Multi-Agent Evaluation}

\author{Mufeng Yang}
\authornote{Mufeng Yang and Junwei Yu contributed equally to the paper.}
\orcid{0009-0005-1714-2544}
\affiliation{%
  \institution{University of Tsukuba}
  \city{Ibaraki}
  \country{Japan}
}
\email{s2321728@u.tsukuba.ac.jp}

\author{Junwei Yu}
\authornotemark[1]
\orcid{0009-0004-1657-3310} 
\affiliation{%
  \institution{The University of Tokyo}
  \city{Tokyo}
  \country{Japan}
}
\email{yujw@satolab.itc.u-tokyo.ac.jp}

\author{Yepeng Ding}
\correspondingauthor
\orcid{0000-0002-6996-9333}
\affiliation{%
  \institution{Hiroshima University}
  \city{Higashihiroshima}
  \country{Japan}
}
\email{ypding@hiroshima-u.ac.jp}


\begin{abstract}
Large language models (LLMs) are increasingly deployed as automated judges for AI-generated content, yet a single judge is unreliable and even a panel of judges leaves a hard residue: when judges disagree, majority voting simply discards the conflict instead of resolving it. We present JuryFlow, a disagreement-guided, human-in-the-loop multi-agent evaluation framework that treats inter-judge disagreement not as noise to be averaged away, but as a precise, claim-level signal indicating where an evaluation is uncertain. JuryFlow decomposes each candidate response into atomic claims, has a panel of heterogeneous judges assign per-claim verdicts, and builds a disagreement graph whose nodes are scored by verdict entropy and whose edges encode structural similarity between claims. A human acts as a structural guide, i.e., selecting which disagreement to resolve through a single, minimal intervention rather than re-labeling the response, after which the focal claim is re-evaluated, the correction propagates along graph edges and to historically similar cases, and is crystallized into reusable rubric entries that all judges inherit, making the evaluator progressively self-refining. To enable large-scale, reproducible benchmarking without human studies, we evaluate JuryFlow in an automatic configuration in which the focal selection is made by entropy ranking. On MT-Bench and LLMBar, JuryFlow improves agreement with gold labels over single-judge and majority-vote panel baselines, and ablations isolate the contributions of disagreement-targeted re-evaluation, propagation, and rubric induction. We contribute (1) a human-in-the-loop paradigm that recasts the human from labeler to structural guide, (2) the JuryFlow framework operationalizing it through a disagreement graph, focal re-evaluation, and closed-loop rubric induction, and (3) an evaluation protocol with ablations that isolate where the gains originate.
\end{abstract}


\begin{teaserfigure}
  \centering
  \includegraphics[width=0.58\textwidth]{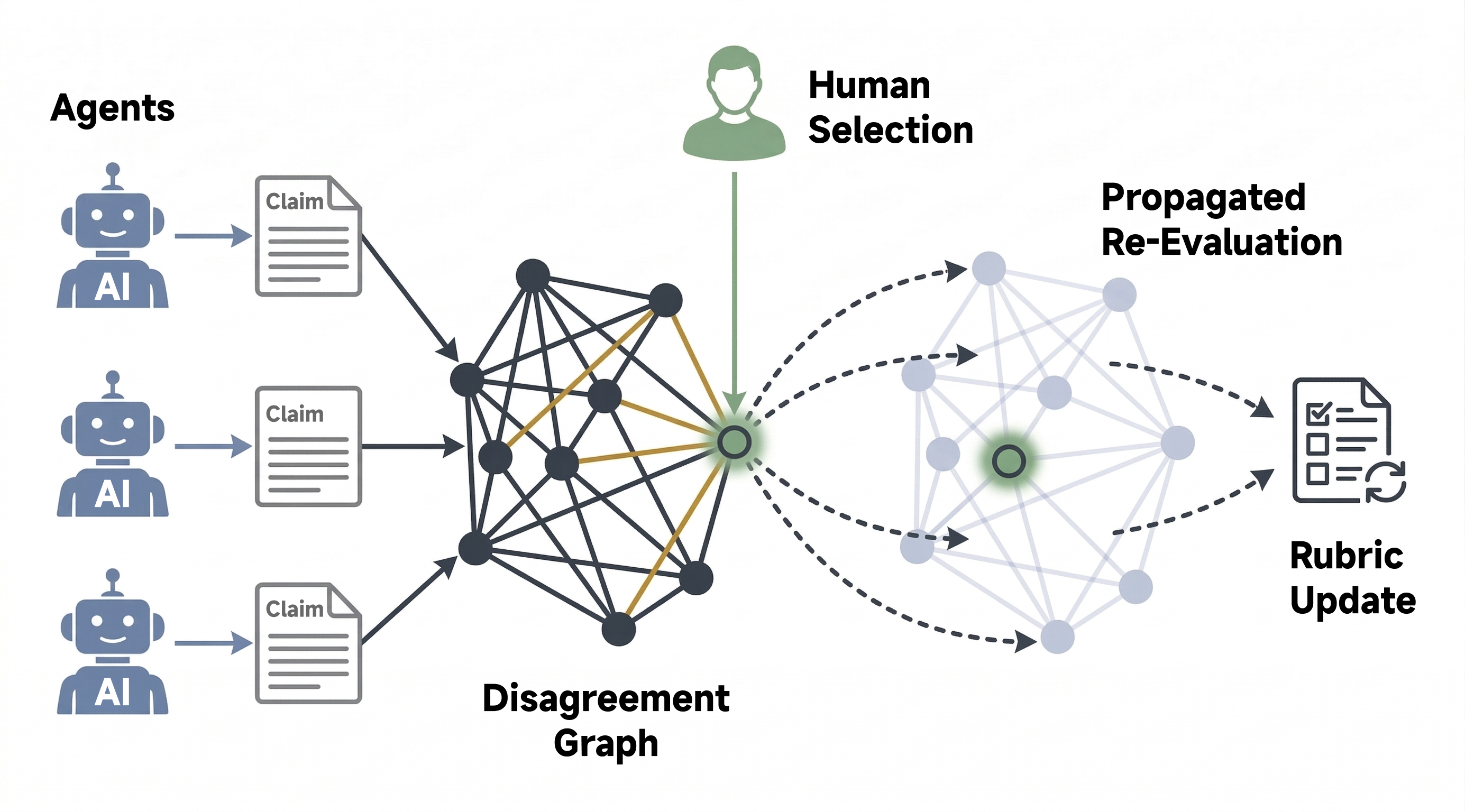}
  \caption{Conceptual overview of JuryFlow. Multiple LLM \emph{Agents} extract \emph{Claims} from a candidate response and produce per-claim verdicts, forming a \emph{Disagreement Graph} that surfaces conflicts. A \emph{Human} selects the most critical disagreement point---the only human intervention (made automatically by entropy ranking in our experiments). The system then performs \emph{Propagated Re-Evaluation} across related claims and historical cases, and abstracts the correction into a \emph{Rubric Update} that all agents inherit, closing the loop.}
  \Description{A conceptual illustration showing the JuryFlow pipeline from left to right. Three AI robot icons (labeled Agents) each produce a document icon (labeled Claim). Arrows flow from agents and claims into a central disagreement graph, a dense network of interconnected nodes with some edges highlighted in amber to indicate conflict. A human selects the highest-entropy focal node (automated in our experiments). Dashed arrows lead rightward to a second, lighter graph representing Propagated Re-Evaluation, and finally to a checklist document icon labeled Rubric Update.}
  \label{fig:teaser}
\end{teaserfigure}


\maketitle

\section{Introduction}
Large language models (LLMs) are increasingly used as automated judges to
evaluate AI-generated content at scale, such as ranking chatbot responses,
scoring factuality, and screening for safety~\cite{zheng2023judging, liu2023geval,
dubois2024alpacafarm}.
A single LLM judge is susceptible to position, verbosity, and self-enhancement
biases~\cite{wang2023large, zheng2024llm, panickssery2024selfpreference}.
Its verdicts are particularly fragile in fine-grained, few-shot settings with
ambiguous criteria, conflicting standards, or incomplete evidence. A
\emph{panel} of heterogeneous judges can reduce model-specific bias and improve
agreement with human preferences~\cite{verga2024poll, chan2023chateval}. Yet
the panel still requires a resolution procedure when its judges disagree.

The standard response is to aggregate the conflicting verdicts through
majority voting or confidence-weighted reconciliation~\cite{chen2024reconcile,
jiang2023llm}.
This procedure discards useful information. Inter-judge disagreement provides
a claim-level indication of \emph{where} an evaluation is
uncertain~\cite{seung1992query, lakshminarayanan2017simple}. Claims with
conflicting verdicts are also those for which a holistic verdict is least
reliable. Existing methods either operate at a coarser granularity or require
more human effort. Uncertainty-driven escalation routes entire uncertain
\emph{instances} to
stronger models or human annotators~\cite{jung2025trust, wang2024lapras,
kim2024meganno}. These systems re-evaluate a full response rather than the
contested claim within it. Interactive criteria tools such as EvalGen and
MetricMate instead address the upstream task of specifying what to evaluate
before inspecting a case~\cite{shankar2024evalgen,
2025_MetricMate-Interactive-Tool_Gebreegziabher}. A complementary mechanism is
needed to locate contested claims automatically, direct additional evaluation
to them, and retain the resulting corrections for subsequent evaluations.

We therefore treat disagreement as an evaluation signal and assign the human
the role of \emph{structural guide} rather than \emph{labeler}. The human makes
one selection instead of re-evaluating the full response. The system then
performs an evidence-based re-evaluation of the selected claim. A structured
\emph{disagreement graph} exposes conflicts, and verdict entropy ranks claims
for selection. The human therefore chooses from a small set of claims with high
entropy. The same ranking supports automatic selection when no human is
available, as in our large-scale evaluation.

We implement this approach in \textbf{JuryFlow}, a disagreement-guided,
human-in-the-loop evaluation framework that operates in five stages
(Figure~\ref{fig:teaser}).
(1)~A panel of heterogeneous judge agents independently decomposes a candidate
response into atomic claims and assigns per-claim verdicts and dimension tags.
(2)~The system constructs a disagreement graph whose nodes carry entropy-based
disagreement scores and whose edges encode structural similarity between claims.
(3)~A human makes the only structural intervention by selecting the focal claim
to resolve. For automated benchmarking, the system instead selects claims with
the highest entropy using the same ranking.
(4)~Each focal selection triggers \emph{propagated re-evaluation}: the focal
claim is re-judged against evidence, and the correction propagates along graph
edges to structurally related claims and, via embedding retrieval, to
historically similar cases.
(5)~The correction is abstracted into a reusable rubric entry that all judges
inherit in subsequent evaluations.
Stages~4 and~5 extend each targeted correction to related claims and incorporate
it into subsequent evaluation criteria. JuryFlow can therefore improve across
instances, unlike panels that evaluate each instance independently.

We evaluate JuryFlow on the MT-Bench and LLMBar
benchmarks~\cite{zheng2023judging, zeng2024llmbar} against single-judge and
majority-vote panel baselines. Because human studies are out of scope here, we
use the automatic-selection configuration, which isolates the multi-agent
machinery from human factors.
We therefore use \emph{human-in-the-loop} throughout to name the interaction
design rather than an empirically validated property of this study. The human's
only action is to select a claim in the disagreement graph, not to assign a
label or rewrite a verdict. Every reported result uses entropy ranking in place
of that selection. Section~\ref{sec:discussion} describes the information that
a human could provide beyond this proxy, and
Section~\ref{sec:limitations} lists the unvalidated mode as a limitation.
Four questions organize the experiments: whether
disagreement-targeted re-evaluation beats aggregation, how the gain splits
between propagation (Stage~4) and rubric induction (Stage~5), how the
trade-off between accuracy and cost varies with the propagation threshold, and
whether the gains require genuine model diversity rather than ensemble size alone
(Section~\ref{sec:results}).

This paper makes three contributions.
\textbf{(1)~A formulation of disagreement as an actionable signal:} we use
inter-judge disagreement to identify claims that require additional inference,
replacing holistic aggregation with targeted re-evaluation.
\textbf{(2)~The JuryFlow framework}, which operationalizes that signal through a
disagreement graph, human-guided focal selection (automated by entropy ranking
for unattended evaluation), graph- and retrieval-based correction propagation,
and incremental rubric induction that links disagreement resolution to criteria
refinement.
\textbf{(3)~An evaluation protocol with ablations} on MT-Bench and LLMBar that
isolates the contributions of focal re-evaluation, propagation, and rubric
induction and characterizes the trade-off between accuracy and cost.

\section{Related Work}

The following subsections relate JuryFlow to work on LLM-based evaluation,
claim-level decomposition, iterative evaluation alignment, uncertainty in
multi-agent systems, and human-AI collaboration.

\subsection{LLM-as-a-Judge and Multi-Agent Evaluation}

LLMs are widely used to evaluate generated content~\cite{zheng2023judging,
liu2023geval, dubois2024alpacafarm}. Zheng et
al.~\cite{zheng2023judging} introduced MT-Bench and found that GPT-4 judgments
correlate strongly with human preferences. Single-model judges, however, exhibit
position, verbosity, and self-enhancement biases~\cite{wang2023large,
zheng2024llm}.
Multi-agent frameworks attempt to reduce these biases. ChatEval uses role-playing
agents to create prompt-level persona diversity~\cite{chan2023chateval}; ReConcile
resolves disagreement by multi-round discussion and confidence-weighted voting
across genuinely different models rather than prompt variation
alone~\cite{chen2024reconcile}; and Auto-J adapts evaluation criteria to each
scenario~\cite{li2023generative}.
Verga et al.~\cite{verga2024poll} showed that a \emph{panel of
diverse models} (PoLL) can match or exceed a single large judge while reducing
model-specific bias. This result motivates the panel design in JuryFlow and is
consistent with evidence that judges favor their own generations over equally
good alternatives~\cite{panickssery2024selfpreference}. Zeng et
al.~\cite{zeng2024llmbar} introduced LLMBar, a benchmark containing cases that
induce disagreement among judges. These approaches resolve disagreement through
aggregation or majority voting. JuryFlow instead represents disagreement as a
structured signal for targeted intervention.

\subsection{Claim-Level Decomposition and Iterative Evaluation Alignment}

Fine-grained evaluation decomposes responses into verifiable units rather than
assigning a holistic verdict. For example, \citet{min2023factscore} use atomic
facts as the unit of factuality evaluation and show that binary holistic
judgments are inadequate when a response contains both supported and unsupported claims;
\citet{wei2024safe} add search-augmented verification against retrieved
evidence; and \citet{kim2024prometheus} carry claim-level analysis into
open-ended evaluation through customizable rubrics.
Evaluation design also varies by domain. In particular, \citet{ye2023flask}
decompose evaluation into 12 fine-grained skills with separate rubrics and show that
dimension-specific prompting outperforms unified assessment. Safety evaluation
requires judge architectures that differ from those used for factuality or
coherence~\cite{inan2023llamaguard}.

Beyond static rubrics, iterative alignment between human intent and automated
evaluation has been studied through mixed-initiative interfaces.
\citet{shankar2024evalgen} define \emph{criteria drift} as the change in users'
evaluation standards as they inspect more outputs. They identify this drift as
a central obstacle to specifying criteria before observing model behavior.
\citet{kim2024evallm} show that interactive criteria refinement reduces the
number of required prompt revisions by 59\%. In addition,
\citet{2025_MetricMate-Interactive-Tool_Gebreegziabher} present MetricMate,
which supports hierarchical criteria definition and calibration through curated
success/failure examples.
These tools address the upstream task of articulating and stabilizing evaluation
criteria. JuryFlow addresses the complementary downstream problem of resolving
claim-level disagreement under established criteria. It identifies the claim
with the highest entropy within an instance and directs re-evaluation to that
claim.

\subsection{Uncertainty-Driven Human Escalation and Active Learning}

A related line of work studies when and how uncertain automated judgments should
be escalated for human oversight.
\citet{jung2025trust} propose \emph{Trust or Escalate}, establishing provable
guarantees of human agreement by routing low-confidence LLM evaluations to
stronger models through a cascaded selective evaluation framework;
\citet{wang2024lapras} show that routing only low-verifier-score labels to
human re-annotators cuts annotation cost while preserving label quality; and
\citet{kim2024meganno} operationalize the principle in \textsc{MEGAnno+}, a
deployed system for human verification of LLM-generated labels.
Selective routing is related to active learning~\cite{settles2009active}, particularly
query-by-committee~\cite{seung1992query}, which uses disagreement among models
to identify uncertain instances.
These approaches frame escalation at the \emph{instance} level by selecting
which outputs require human attention. JuryFlow operates at the claim level and
does not escalate the judgment to a human. Given an uncertain instance, it
identifies the claim that requires focused re-evaluation and replaces full
response review with targeted computation.

\subsection{Disagreement and Uncertainty in Multi-Agent Systems}

Disagreement among models is a recognized signal of uncertainty in ensemble
methods~\cite{lakshminarayanan2017simple} and committee-based active
learning~\cite{seung1992query}, and in NLP it has been used to calibrate
confidence estimates~\cite{xiong2023can}.
In LLM contexts, \citet{jiang2023llm} use self-consistency and voting across
multiple generations, while
\citet{2019_Principles-Limits-AlgorithmintheLoop_Green} show that people
struggle to calibrate their reliance on algorithmic advice even when
disagreement signals are available.
These approaches treat disagreement as a scalar quantity for aggregation. The
JuryFlow disagreement graph instead represents conflict at the claim level, so
the system can direct re-evaluation to specific claims.

\subsection{Human-AI Collaboration and Mixed-Initiative Interaction}

JuryFlow follows the mixed-initiative interaction
paradigm~\cite{1999_Principles-Mixedinitiative-User_Horvitz}, which allocates
control between people and machines according to their respective strengths.
Its design also draws on empirical studies of human-AI decision
making~\cite{2023_Science-HumanAI-Decision_Lai}.
Because people overrely on AI when verification is
costly~\cite{2021_Trust-Think-Cognitive_Bucinca,
2023_Explanations-Can-Reduce_Vasconcelos}, we keep the human's action minimal: a
single structural selection over a disagreement graph rather than a holistic
re-evaluation. This design reduces the cost of oversight while leaving the
final verdict to the system.
Zeno~\cite{2023_Zeno-Interactive-Framework_Cabrera} and
ChainForge~\cite{2024_ChainForge-Visual-Toolkit_Arawjo} support human-directed
analysis at the \emph{dataset} level for behavioral evaluation and visual prompt
engineering, respectively. JuryFlow instead organizes within-instance
disagreement so that a single selection, made either by a human or by entropy
ranking, determines the target of re-evaluation.

\section{Method}

\begin{figure*}[t]
  \includegraphics[width=\textwidth]{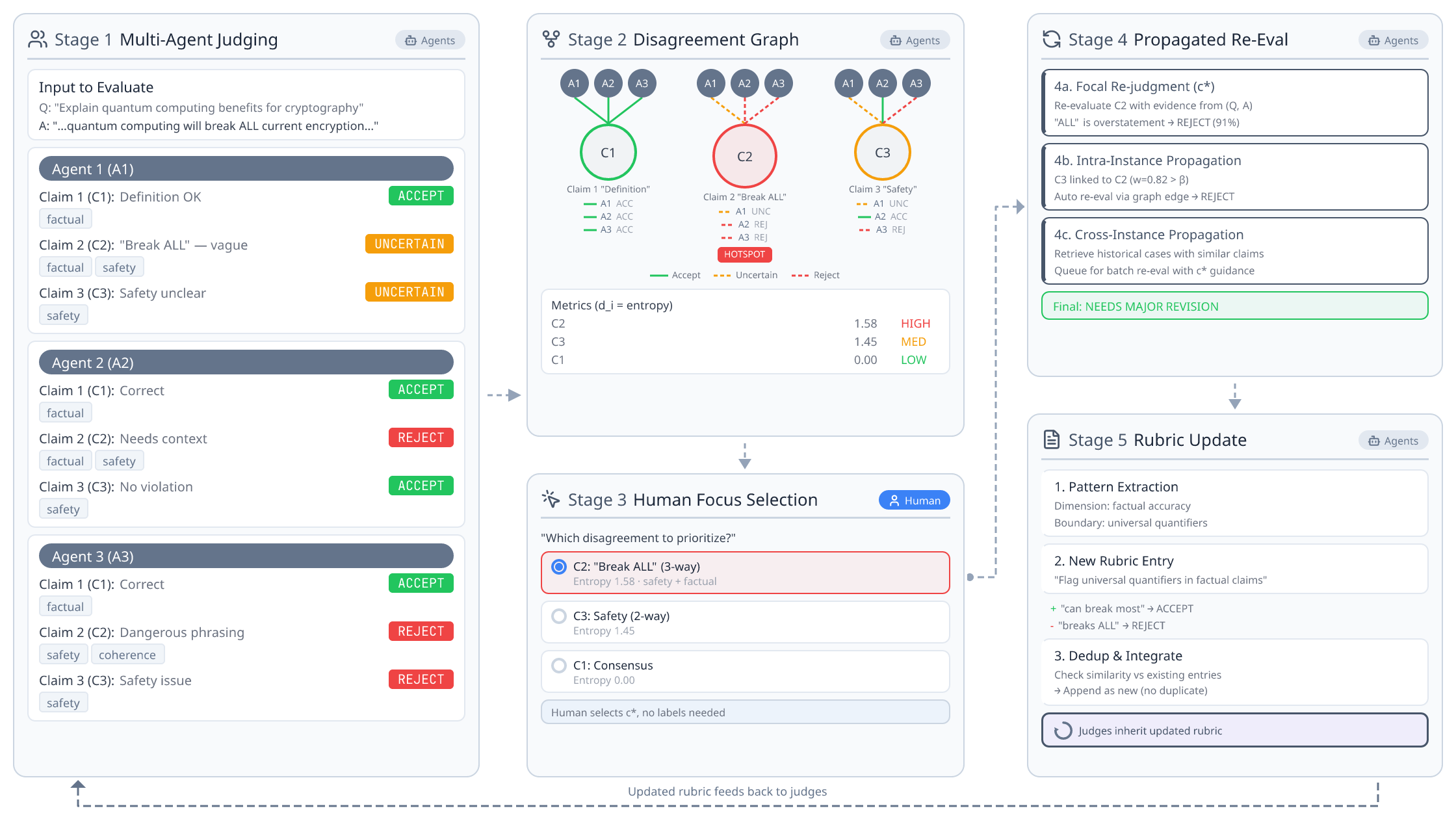}
  \caption{The five-stage JuryFlow pipeline, illustrated with a cryptography question whose candidate response contains three claims: C1 (``Definition''), C2 (``Break ALL current encryption''), and C3 (``Safety''). Each stage is labeled \emph{Agents} (automated) or \emph{Human}. Stage~1 produces per-claim verdicts and dimension tags. Stage~2 constructs the disagreement graph and detects high-entropy claims. Stage~3 contains the only human input, a single focal-claim selection. Stage~4 propagates corrections through graph edges and cross-instance retrieval. Stage~5 converts the correction into a reusable rubric entry.}
  \Description{A three-column workflow diagram showing five stages of the JuryFlow evaluation pipeline. Stage 3 carries a Human badge, the others an Agents badge. Column 1 (Stage 1) displays three AI judge agents with per-claim verdicts and tags. Column 2 has Stage 2 (disagreement graph with claim nodes and verdict-colored edges) above Stage 3 (human focal selection). An elbow connector leads to Column 3 with Stage 4 (focal re-judgment, intra/cross-instance propagation, final verdict) above Stage 5 (pattern extraction, rubric entry generation, deduplication). A dashed polyline arrow along the bottom connects Stage 5 back to Stage 1.}
  \label{fig:pipeline}
\end{figure*}

Unlike methods that aggregate judgments by voting or treat human input as an
isolated correction, \textbf{JuryFlow} uses each selected disagreement to update
structurally similar cases and reusable evaluation criteria. This process links
downstream disagreement resolution to subsequent criteria refinement. The
framework has five stages, as shown in Figure~\ref{fig:pipeline}.

\subsection{Stage 1: Multi-Agent Structured Judging}

Given an evaluation instance consisting of a prompt $Q$ and a candidate
response $A$, we invoke a panel of $N$ heterogeneous judge agents
$\{J_1, J_2, \dots, J_N\}$.
We obtain \emph{model diversity} primarily by using LLMs from different
providers, which have distinct training data and architectural
biases~\cite{verga2024poll}. We optionally add \emph{prompt diversity} by
varying evaluation personas or domain-specific
instructions~\cite{chan2023chateval, ye2023flask}. A predefined output schema
enforces consistent claim-level decomposition across providers and makes their
outputs directly comparable.
Each agent independently produces a judgment comprising:
\begin{itemize}
    \item A set of extracted claims $C = \{c_1, c_2, \dots, c_k\}$, where $k$
          is the number of claims and each claim is an atomic, verifiable
          proposition derived from $A$;
    \item A verdict $v_i^{(j)} \in \{\texttt{accept}, \texttt{reject},
          \texttt{uncertain}\}$ assigned by agent $J_j$ to claim $c_i$;
    \item A set of tags $T_i^{(j)} \subseteq \mathcal{T}$, where
          $\mathcal{T}$ is a predefined set of high-level evaluation
          dimensions (e.g., \textit{factual accuracy}, \textit{logical
          coherence}, \textit{safety}).
          Following the fine-grained skill taxonomy of
          \citet{ye2023flask}, we take $\mathcal{T}$ to be a fixed,
          domain-configurable set of 5 to 7 dimensions, which keeps tag
          assignment consistent across agents and stabilizes the
          tag-based similarity of Stage~2.
\end{itemize}
No aggregation or voting occurs at this stage. The disagreement analysis
receives every judgment and its associated tags.

\subsection{Stage 2: Disagreement Graph Construction}
\label{sec:graph}

\begin{figure}[t]
  \centering
  \includegraphics[width=0.85\columnwidth]{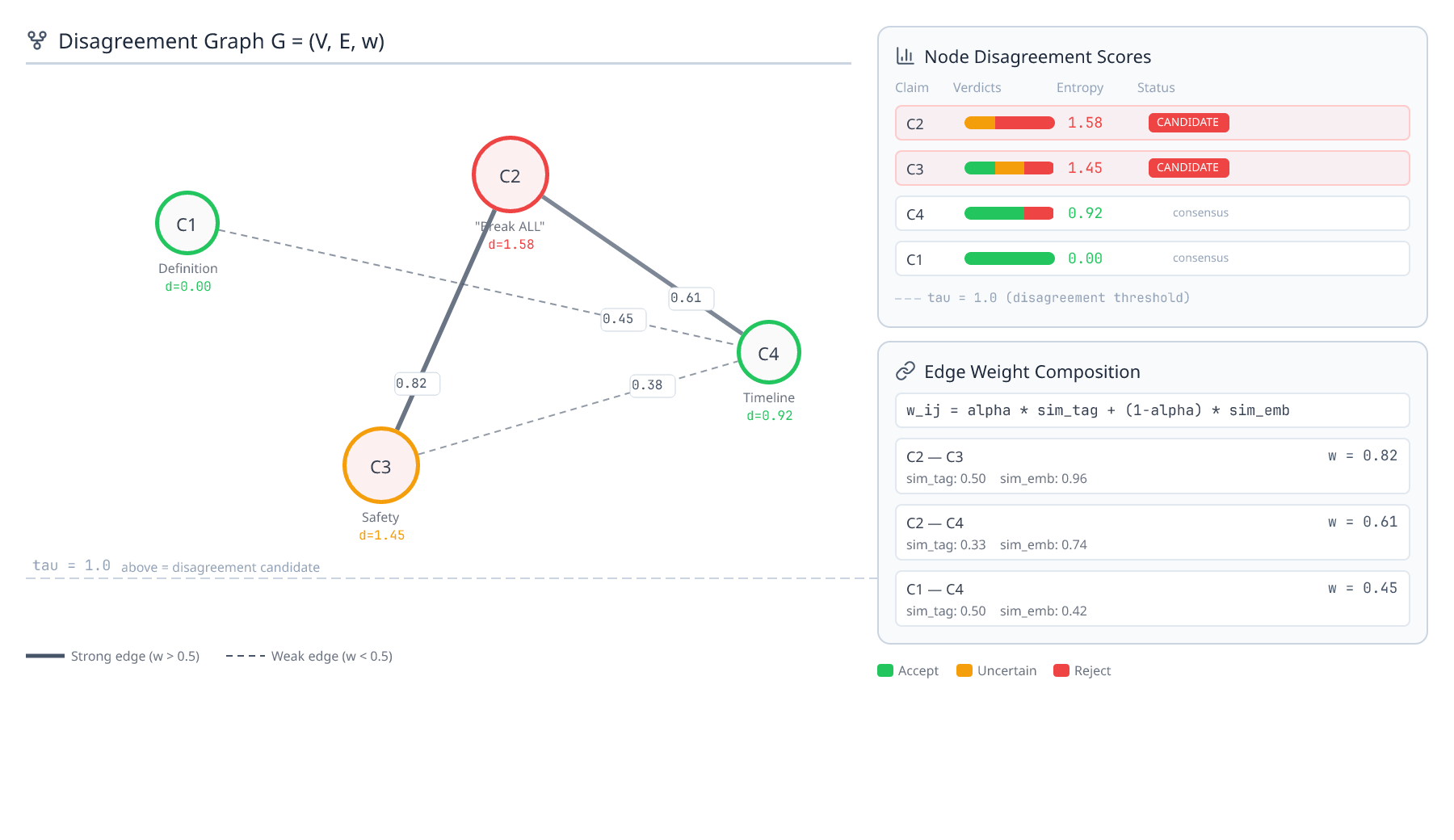}
  \caption{Disagreement graph detail for a four-claim instance. Nodes carry entropy-based disagreement scores $d_i$; those above $\tau$ (dashed line) are disagreement candidates. Edge thickness and opacity encode $w_{ij}$, a mixture of tag Jaccard and embedding cosine similarity. Right: per-claim verdict distributions and the edge-weight decomposition for the three strongest edges.}
  \Description{A two-panel figure. The left panel shows a graph with four claim nodes (C1 green, C2 red, C3 yellow, C4 yellow) connected by edges of varying thickness. C2 and C3 are above the tau threshold line and marked as candidates. Edge labels show w values (0.82, 0.61, 0.45, 0.38). The right panel contains a table of node disagreement scores with stacked verdict bars and a breakdown of edge weight composition showing sim\_tag and sim\_emb values.}
  \label{fig:graph}
\end{figure}

We represent the multi-agent evaluation output as a weighted
\textbf{disagreement graph} $G = (V, E, w)$, where each node $v_i \in V$
corresponds to a claim $c_i$ and each edge $e_{ij} \in E$ encodes the
structural relationship between two claims. Figure~\ref{fig:graph} shows a
worked four-claim instance.

\paragraph{Node-level disagreement score.}
For each claim $c_i$, we compute a disagreement score $d_i$ based on the
entropy of the verdict distribution across agents:
\begin{equation}
    d_i = -\sum_{v \in \mathcal{V}} p_i(v) \log p_i(v),
\end{equation}
where $p_i(v)$ is the proportion of agents assigning verdict $v$ to claim
$c_i$, and $\mathcal{V} = \{\texttt{accept}, \texttt{reject},
\texttt{uncertain}\}$.
Claims with $d_i$ exceeding a threshold $\tau$ are marked as
\textit{disagreement candidates}.

\paragraph{Edge weights.}
The weight $w_{ij}$ between two claim nodes captures their structural
similarity along two complementary dimensions:
\begin{equation}
    w_{ij} = \alpha \cdot \text{sim}_{\text{tag}}(c_i, c_j)
            + (1 - \alpha) \cdot \text{sim}_{\text{emb}}(c_i, c_j),
\end{equation}
where $\text{sim}_{\text{tag}}$ is the Jaccard similarity over the
union of agent-assigned tags, $\text{sim}_{\text{emb}}$ is the cosine
similarity of claim embeddings, and $\alpha \in [0, 1]$ is a mixing
coefficient.
The two terms capture distinct relationships between a pair of claims.
$\text{sim}_{\text{tag}}$ measures whether the panel evaluated the claims along
the same dimensions, whereas $\text{sim}_{\text{emb}}$ measures whether the
claims express similar content. A factual error and a safety error in the same
sentence score high on the first measure and low on the second. Two paraphrases
of one assertion exhibit the reverse pattern. A correction can transfer when
either relationship holds, so we use a convex combination that is monotonic in
both signals. A product would instead require both forms of similarity. The
coefficient $\alpha$ controls their relative contributions.
Because $\mathcal{T}$ is a fixed, finite set shared by all agents,
$\text{sim}_{\text{tag}}$ is well-defined and bounded. Agents may
disagree on \emph{verdicts} while still agreeing on which
\emph{dimensions} are relevant to a claim, a similarity signal
independent of verdict agreement.
The graph identifies both contentious claims, which have high $d_i$, and
structurally related claims, which have high $w_{ij}$. Stage~4 uses this
structure to propagate focal corrections.

\subsection{Stage 3: Human-Guided Focal Selection}

\begin{figure}[t]
  \centering
  \includegraphics[width=0.85\columnwidth]{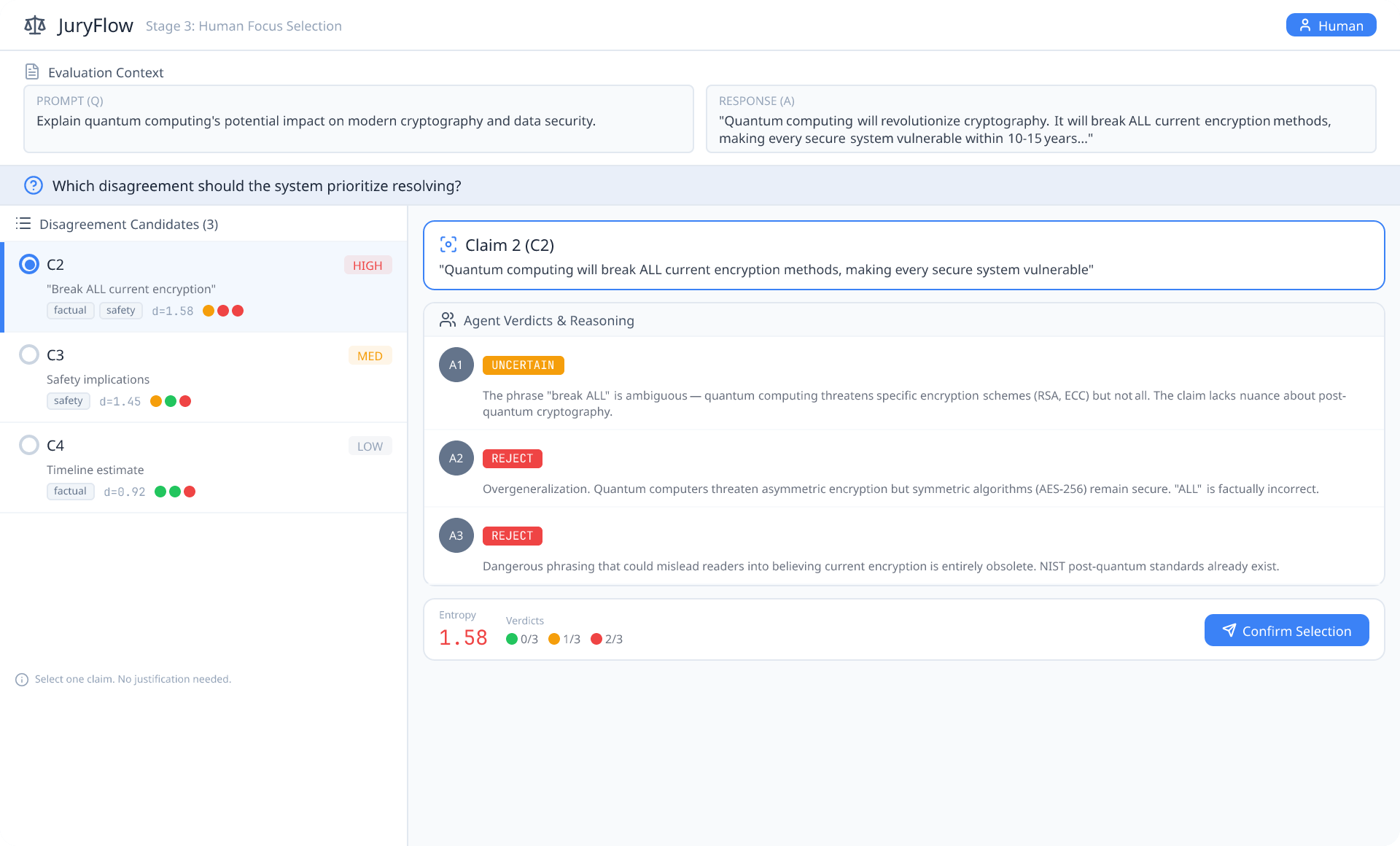}
  \caption{Human-guided focal selection interface. Left: disagreement candidates ranked by entropy $d_i$. Right: the selected claim's per-agent verdicts and rationales. The human confirms a single selection with no justification required; in our experiments the selection is made automatically by entropy ranking (Section~\ref{sec:setup}).}
  \Description{A web application mockup with a Human Review Mode badge. The left sidebar lists three disagreement candidates with entropy values and dimension tags; one is selected via a radio button. The right detail panel shows the selected claim's per-agent verdicts with reasons, an entropy score, and a Confirm Selection button.}
  \label{fig:ui}
\end{figure}

JuryFlow is designed as a human-in-the-loop framework. Let $m$ denote the focal
budget for an instance. The system presents the disagreement graph and
highlights the top-$m$ candidates ranked by $d_i$ (Figure~\ref{fig:ui}). It then
asks the human one question:
\emph{which disagreement should the system resolve first?}
The human selects one focal claim $c^*$; no verdict, label, or justification is
required. This selection is the only human input. The system remains responsible
for the final verdict.

For large-scale, reproducible benchmarking without human studies, JuryFlow also
supports an \emph{automatic} focal-selection mode: the system selects the
top-$m$ claims whose score exceeds $\tau$, with $c^{*} = \arg\max_{i} d_i$
anchoring propagation in Stage~4 and the remaining candidates re-evaluated
independently.
Verdict entropy is maximal when the panel is most divided. Thus, $d_i$ provides
a reproducible proxy for human selection by locating claims for which an
aggregated verdict is least reliable.
The budget $m$ and threshold $\tau$ trade re-evaluation cost against coverage.
All experiments in this paper use the automatic mode (Section~\ref{sec:setup}),
isolating the multi-agent machinery from human factors; a user study of the
human-guided mode is left to future work.

\subsection{Stage 4: Propagated Re-Evaluation}
\label{sec:propagation}

\begin{figure}[t]
  \centering
  \includegraphics[width=0.85\columnwidth]{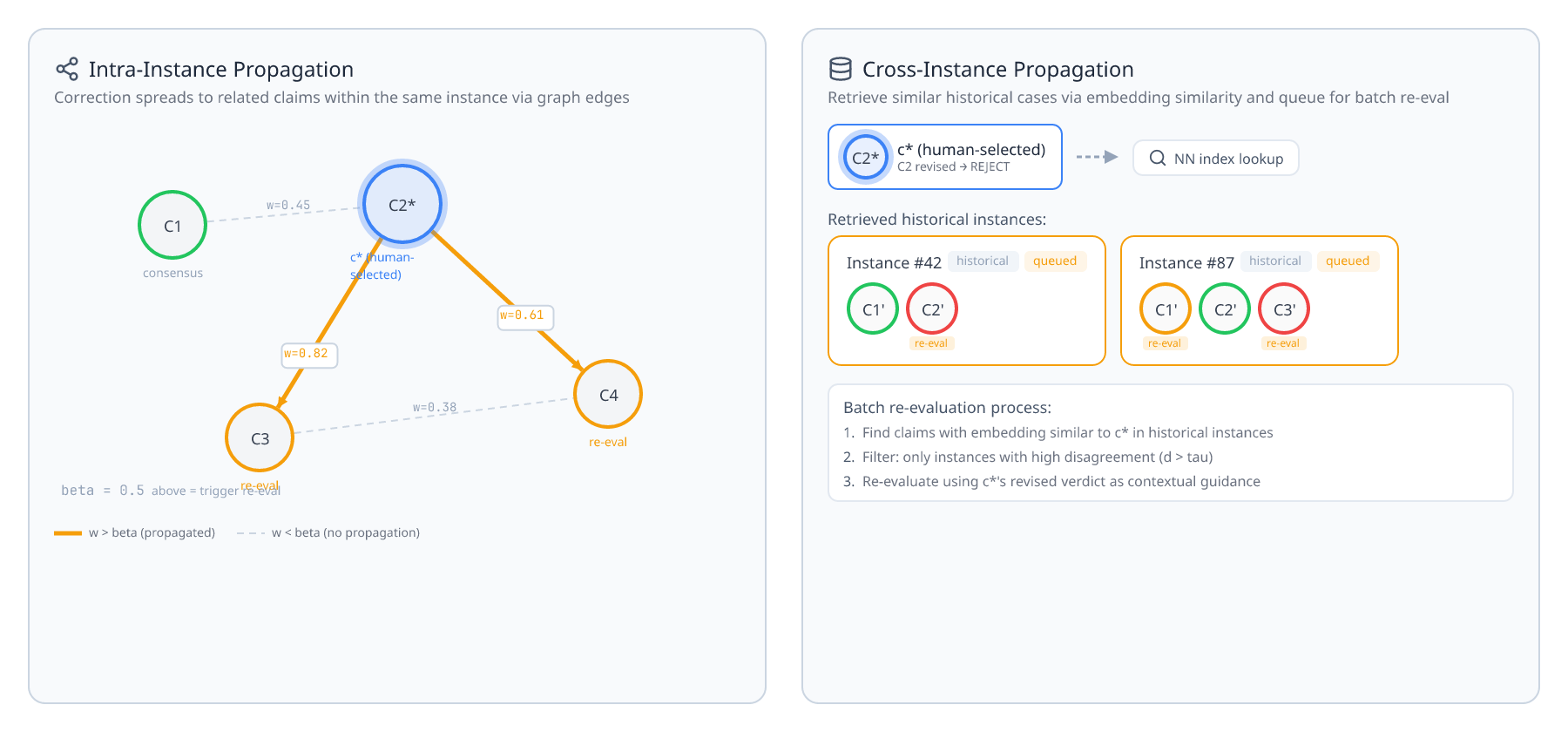}
  \caption{Two propagation mechanisms triggered by selecting $c^*$. \textbf{Left:} Intra-instance propagation flags claims connected to $c^*$ by edges with $w > \beta$ for re-evaluation (amber); claims connected by edges below the threshold (dashed gray) are unaffected. \textbf{Right:} Cross-instance propagation uses nearest-neighbor embedding search to retrieve historical instances with similar, high-disagreement claims. These claims are queued for batch re-evaluation using the revised verdict for $c^*$ as guidance.}
  \Description{A two-panel figure. Left panel (Intra-Instance Propagation) shows a graph with focal claim C2 (blue, selected) connected by amber arrows to C3 and C4 (both marked re-eval), with edge weights 0.82 and 0.61 above beta=0.5. C1 is connected by dashed gray lines (below beta). Right panel (Cross-Instance Propagation) shows the focal claim C2 feeding into a nearest-neighbor index lookup, which retrieves two historical instances (42 and 87) each containing claims marked for re-evaluation. A batch re-evaluation process lists three steps: find similar claims, filter by disagreement, and re-evaluate with revised verdict as guidance.}
  \label{fig:propagation}
\end{figure}

Selecting $c^*$ triggers the two re-evaluation processes shown in
Figure~\ref{fig:propagation}.

\paragraph{Focal re-judgment.}
A reviewer agent re-evaluates the focal claim $c^*$ using supporting and
opposing evidence from the original context $(Q, A)$. It returns a revised
verdict $v^*$ with an evidence-grounded rationale (Appendix~\ref{app:prompts}).

\paragraph{Graph-based propagation.}
The correction is then propagated through $G$ in two directions. Let $\beta$
denote the propagation threshold.
\emph{Intra-instance:} within the current instance, claims $c_j$ with edge
weight $w_{c^* j} > \beta$ are flagged for automatic re-evaluation by the
reviewer agent, on the assumption that strongly related claims are affected by
the same underlying issue.
\emph{Cross-instance:} the system retrieves historical instances whose claim
embeddings are similar to $c^*$ using a nearest-neighbor index over all
previously evaluated claims. Historical instances that contain
high-disagreement claims are queued for batch re-evaluation. The reviewer uses
the revised verdict and rationale for $c^*$ as contextual guidance. The
threshold $\beta$ controls the trade-off between correction coverage and
computational cost.

\paragraph{Final verdict.}
For each evaluated or re-evaluated instance, the system combines the original
agent judgments with the focused re-judgments. Evidence-grounded revisions of
high-disagreement claims receive priority in the final verdict.

\subsection{Stage 5: Incremental Rubric Update}
\label{sec:rubric}

Each focal correction encodes an implicit evaluation preference that may apply
beyond the current instance. JuryFlow converts this preference into a reusable
rubric entry in three steps. During \emph{pattern extraction}, the system
analyzes $c^*$, its tags, the original conflicting verdicts, and the revised
verdict. This analysis identifies the evaluation dimension and boundary
condition associated with the disagreement. During \emph{rubric entry
generation}, an LLM produces a candidate entry containing a natural-language
criterion, one positive example from the corrected case, and one negative
example from the original erroneous judgment. During \emph{deduplication and
integration}, the system compares the candidate with existing entries using
embedding similarity. It adds the new example to a near-duplicate entry or
appends the candidate when no near-duplicate exists.

All judge agents inherit the updated rubric in subsequent evaluations. The
system can therefore accumulate corrections and reduce residual disagreement on
recurring patterns. One selection in Stage~3 can affect related claims through
Stage~4 and future evaluation criteria through Stage~5. Disagreement is thus
retained for targeted re-evaluation, and downstream corrections inform later
criteria.

\section{Experimental Setup}
\label{sec:setup}

JuryFlow is a human-in-the-loop framework, but to enable large-scale,
reproducible benchmarking we evaluate its \emph{automatic-selection}
configuration, in which the focal claim (Stage~3) is chosen by entropy ranking
rather than by a human. This configuration evaluates the multi-agent panel,
disagreement graph, propagation, and rubric induction without introducing human
factors. We leave evaluation of the human-guided configuration to future work.
All results are averaged over three random seeds and reported as
mean~$\pm$~standard deviation where applicable.

\subsection{Datasets}
We use two complementary public benchmarks.
\textbf{MT-Bench}~\cite{zheng2023judging} contains multi-turn questions spanning
reasoning, writing, math, and knowledge, with human preference annotations
serving as gold labels; we evaluate $300$ pairwise-preference instances.
\textbf{LLMBar}~\cite{zeng2024llmbar} is a meta-evaluation benchmark whose
instances pair two responses with a single objectively better option; its
\textsc{Natural} and \textsc{Adversarial} splits contain cases prone to judge
disagreement. We evaluate all $419$ instances ($100$ \textsc{Natural} and $319$
\textsc{Adversarial}: \textsc{Neighbor}~134, \textsc{GPTInst}~92,
\textsc{GPTOut}~47, \textsc{Manual}~46). Gold labels come directly from the
datasets' released human annotations; we add no manual labeling.

\subsection{Pairwise Evaluation Protocol}
Both benchmarks use pairwise comparisons: each instance contains two candidate
responses and asks which is preferable. In contrast, the JuryFlow pipeline from
Section~\ref{sec:graph} onward evaluates one response. We therefore apply the
full pipeline independently to each response. Each response is decomposed into
claims, evaluated by the panel, and subjected to propagated re-evaluation for
claims with high disagreement.
Each response's final per-claim verdicts are reduced to a response-level quality
score using a weakest-link rule. This rule penalizes a response in proportion to
its rejected and uncertain claims. JuryFlow predicts that the response with the
higher score is preferable. It resolves ties using the holistic verdict from
the \emph{strongest judge}, defined as the panel member with the highest
standalone accuracy on the development split. This judge also serves as the
single-judge baseline in Section~\ref{sec:baselines}.
Accuracy and Cohen's~$\kappa$ are measured against each dataset's gold preferred
response. Both baselines use the same reduction and omit only
disagreement-guided re-evaluation, thereby isolating its contribution.

\subsection{Judge Panel and Implementation}
The panel comprises $N = 5$ heterogeneous judge agents: GPT-4o, Claude~3.5
Sonnet, Gemini~1.5 Pro, Llama-3.1-70B-Instruct, and
Qwen2.5-72B-Instruct. We obtain heterogeneity primarily from distinct LLM
families~\cite{verga2024poll} and, in the persona condition, from prompt
diversity~\cite{chan2023chateval}. Each agent follows a fixed output schema to
produce claims, per-claim verdicts, and dimension tags
(Appendix~\ref{app:prompts}). Claim embeddings for
$\text{sim}_{\text{emb}}$ and cross-instance retrieval are produced by
\texttt{text-embedding-3-large}. We index them with the FAISS library using a
hierarchical navigable small-world (HNSW) graph for approximate nearest-neighbor
retrieval.
The Stage~4 reviewer is instantiated from the \emph{same} model as the strongest
panel judge, so no model outside the panel is introduced and the gains in
Table~\ref{tab:main} cannot be attributed to a stronger second opinion.
Decoding temperature is fixed to $0.0$ (greedy). Residual variation across seeds
arises from provider-side nondeterminism and instance order. The seed permutes
the instance order, which affects cross-instance retrieval and the sequence in
which rubric entries accumulate. The $\pm$ ranges in Table~\ref{tab:main}
therefore also measure sensitivity to ordering.

\subsection{Baselines}
\label{sec:baselines}
We compare JuryFlow with two aggregation baselines that use the same panel but
omit disagreement-guided re-evaluation. \textbf{Single judge} uses a holistic
verdict from the strongest individual model in the panel, following the
standard LLM-as-a-judge setup. \textbf{Majority-vote panel} aggregates verdicts
from the full panel by majority vote and uses the strongest model to break ties,
following the PoLL-style baseline~\cite{verga2024poll}.
Both baselines use the same models, prompts, and claim decomposition as JuryFlow,
which isolates the effect of disagreement-guided re-evaluation. We omit
baselines based on full human review or human escalation because our evaluation targets the
automatic-selection configuration (Section~\ref{sec:limitations}).

\subsection{Metrics}
We report \textbf{accuracy} (agreement with gold labels) and \textbf{Cohen's
$\kappa$}, which corrects for chance agreement. To quantify cost we report the
mean \textbf{re-evaluation LLM calls per instance}, capturing the
trade-off between accuracy and computation in Stages~3 and~4.

\subsection{Hyperparameters}
The disagreement threshold $\tau$, edge-mixing coefficient $\alpha$, propagation
threshold $\beta$, focal budget $m$, and panel size $N$ are selected by grid
search on a held-out development split. This split contains data from both
datasets and is disjoint from the evaluated instances (Table~\ref{tab:hparams}).
Section~\ref{sec:results} analyzes sensitivity to $\beta$.

\begin{table}[t]
  \caption{Hyperparameters: role, search grid, and selected value (tuned on a held-out development split). For $\alpha$, the selected $0.4$ places $0.6$ weight on embedding similarity and $0.4$ on tag similarity, reflecting that judges may split on verdicts while agreeing on relevant dimensions.}
  \label{tab:hparams}
  \footnotesize
  \setlength{\tabcolsep}{2pt}
  \begin{tabular}{llcc}
    \toprule
    Symbol   & Role                                & Grid & Selected \\
    \midrule
    $\tau$   & disagreement threshold (Stages 2, 3) & $\{0.3,0.5,0.7,0.9\}$ & $\mathbf{0.7}$ \\
    $\alpha$ & tag/embedding edge mix (Stage 2)    & $\{0.0,0.2,0.4,0.6,0.8,1.0\}$ & $\mathbf{0.4}$ \\
    $\beta$  & propagation threshold (Stage 4)     & $\{0.0,0.3,0.5,0.7,0.9\}$ & $\mathbf{0.5}$ \\
    $m$      & focal budget per instance (Stage 3) & $\{1,2,3,5\}$ & $\mathbf{3}$ \\
    $N$      & panel size (Stage 1)                & $\{3,5,7\}$ & $\mathbf{5}$ \\
    \bottomrule
  \end{tabular}
\end{table}

\section{Results}
\label{sec:results}

We organize results around four research questions. All numbers are averaged
over three seeds; standard deviations appear in Table~\ref{tab:main} and are
omitted elsewhere for readability.

\subsection{RQ1: Does disagreement-guided re-evaluation beat aggregation?}
Table~\ref{tab:main} tests whether directing re-evaluation to high-entropy claims
yields higher agreement with gold labels than aggregating the same panel's
verdicts. JuryFlow improves over the best single judge by $5.3$ accuracy points
on MT-Bench ($0.804 \rightarrow 0.857$) and by $10.5$ points on LLMBar
($0.679 \rightarrow 0.784$). Cohen's $\kappa$ increases from $0.59$ to $0.71$
on MT-Bench and from $0.36$ to $0.57$ on LLMBar. JuryFlow also exceeds the
stronger majority-vote panel by $2.9$ and $6.6$ points, respectively. The gain
is larger on LLMBar, which contains adversarial cases prone to judge
disagreement. On MT-Bench, the panel already has higher agreement with human
preferences, leaving less room for improvement.

\begin{table}[t]
  \caption{Main results: agreement with gold labels on MT-Bench and LLMBar (mean~$\pm$~std over three seeds). Best per column in bold.}
  \label{tab:main}
  \footnotesize
  \begin{tabular}{lcccc}
    \toprule
    & \multicolumn{2}{c}{MT-Bench} & \multicolumn{2}{c}{LLMBar} \\
    \cmidrule(lr){2-3}\cmidrule(lr){4-5}
    Method                   & Acc. & $\kappa$ & Acc. & $\kappa$ \\
    \midrule
    Single judge (best)      & $0.804\pm0.011$ & $0.59$ & $0.679\pm0.014$ & $0.36$ \\
    Majority-vote panel      & $0.828\pm0.009$ & $0.64$ & $0.718\pm0.012$ & $0.44$ \\
    \textbf{JuryFlow (full)} & $\mathbf{0.857\pm0.008}$ & $\mathbf{0.71}$ & $\mathbf{0.784\pm0.010}$ & $\mathbf{0.57}$ \\
    \bottomrule
  \end{tabular}
\end{table}

\subsection{RQ2: Where do the gains come from?}
Table~\ref{tab:ablation} ablates the two amplification stages: removing rubric
induction (Stage~5) measures the contribution of accumulated criteria, removing
propagation (Stage~4) measures the contribution of correction transfer, and
removing both retains only focal re-evaluation.
On LLMBar, focal re-evaluation alone already lifts the majority-vote baseline by
$3.0$ points ($0.718 \rightarrow 0.748$). In leave-one-out comparisons with the
full system, propagation contributes $2.8$ points and rubric induction
contributes another $1.2$ points. Their effects are approximately additive
($0.748 + 2.8 + 1.2 \approx 0.784$), and propagation makes the larger
contribution. The cost results show a similar pattern. Of the $3.6$
re-evaluation calls per instance, propagation accounts for approximately $1.7$
and rubric induction for approximately $0.6$, while focal re-evaluation alone
costs $1.3$. Thus, focal re-evaluation explains $3.0$ of the $6.6$ points gained
over the majority-vote panel on LLMBar, while propagation and rubric induction
jointly explain the remaining $3.6$ points.

\begin{table}[t]
  \caption{Component ablation, where \emph{Calls} denotes the mean number of re-evaluation LLM calls per instance (lower is cheaper). Best per column in bold.}
  \label{tab:ablation}
  \footnotesize
  \setlength{\tabcolsep}{3pt}
  \begin{tabular}{lccc}
    \toprule
    Variant                          & MT-Bench Acc. & LLMBar Acc. & Calls \\
    \midrule
    \textbf{JuryFlow (full)}         & $\mathbf{0.857}$ & $\mathbf{0.784}$ & $3.6$ \\
    \quad $-$ rubric (Stage 5)       & $0.851$ & $0.772$ & $3.0$ \\
    \quad $-$ propagation (Stage 4)  & $0.847$ & $0.756$ & $1.9$ \\
    \quad $-$ both (focal only)      & $0.841$ & $0.748$ & $1.3$ \\
    Majority-vote panel (no re-eval) & $0.828$ & $0.718$ & $0$  \\
    \bottomrule
  \end{tabular}
\end{table}

\subsection{RQ3: How does the trade-off between accuracy and cost vary with propagation?}
The propagation threshold $\beta$ controls how aggressively corrections spread
(Section~\ref{sec:propagation}). Table~\ref{tab:beta} sweeps $\beta$, reporting
accuracy and the number of re-evaluation calls per instance. Accuracy follows an
inverted U shape and reaches its maximum of $0.784$ at $\beta = 0.5$, the value
selected in Table~\ref{tab:hparams}. At $\beta = 0.0$, nearly every related claim
triggers propagation. The number of calls increases to $8.7$ per instance, while
accuracy falls to $0.761$ because errors propagate across weakly related claims.
At $\beta = 0.9$, limited propagation yields performance closer to focal-only
re-evaluation, with $0.756$ accuracy at $1.4$ calls. The selected value
$\beta = 0.5$ lies at the knee of the curve.

\begin{table}[t]
  \caption{Sensitivity to the propagation threshold $\beta$ on LLMBar, where \emph{Calls} denotes the mean number of re-evaluation LLM calls per instance. Best accuracy in bold.}
  \label{tab:beta}
  \footnotesize
  \begin{tabular}{lccccc}
    \toprule
    $\beta$ & 0.0 & 0.3 & 0.5 & 0.7 & 0.9 \\
    \midrule
    Acc.  & $0.761$ & $0.779$ & $\mathbf{0.784}$ & $0.773$ & $0.756$ \\
    Calls & $8.7$ & $5.4$ & $3.6$ & $2.1$ & $1.4$ \\
    \bottomrule
  \end{tabular}
\end{table}

\subsection{RQ4: Does judge heterogeneity matter?}
Table~\ref{tab:hetero} compares three panel constructions at a fixed panel size
of $N = 5$: multi-model, multi-persona (one model with varied prompts), and
mixed (both model and prompt diversity). This comparison tests whether the gains depend on model
diversity~\cite{verga2024poll, panickssery2024selfpreference} or only on ensemble
size. On adversarial LLMBar, a
multi-persona panel from a single model reaches only $0.726$, barely above the
$0.718$ majority-vote baseline. Judges that share a model also share its blind
spots, so their disagreement is less informative about model errors. The
multi-model ($0.779$) and mixed ($0.784$) panels recover the full gain. The
mixed panel is our main configuration (Table~\ref{tab:main}).

\begin{table}[t]
  \caption{Effect of judge heterogeneity source (panel size fixed at $N=5$). Best per column in bold.}
  \label{tab:hetero}
  \footnotesize
  \begin{tabular}{lcc}
    \toprule
    Panel construction & MT-Bench Acc. & LLMBar Acc. \\
    \midrule
    Multi-model   & $0.854$ & $0.779$ \\
    Multi-persona & $0.823$ & $0.726$ \\
    \textbf{Mixed}         & $\mathbf{0.857}$ & $\mathbf{0.784}$ \\
    \bottomrule
  \end{tabular}
\end{table}

\section{Discussion}
\label{sec:discussion}

\paragraph{What the human adds that entropy cannot.}
Although all reported gains come from the automatic configuration, entropy and a
human selector use different information. First, entropy measures how strongly
the panel is divided, not the consequence of the disagreement. It may therefore
rank a strong disagreement about style above a weaker disagreement about safety.
A person familiar with the downstream stakes could reverse that order. Second,
entropy is zero when all judges agree. It cannot identify confident, correlated
errors in a homogeneous panel (Section~\ref{sec:limitations}), whereas a human
could select a claim that every judge accepted. A human can also draw on task
context, domain knowledge, and standards that evolve as more outputs are
inspected~\cite{shankar2024evalgen,
2025_MetricMate-Interactive-Tool_Gebreegziabher}. In this sense, our results are
a \emph{lower bound} for the amplification process when driven by the least
costly selector. They do not measure performance when consensus errors can be
identified by a human.

\paragraph{Why disagreement-targeting helps.}
Verdict entropy provides an inexpensive, model-internal measure of uncertainty.
It is maximal when the panel is evenly divided, where an aggregated verdict is
least reliable. Directing a second, evidence-grounded evaluation to these claims
uses additional inference on cases where it can alter the outcome. Uniform
re-evaluation would also spend computation on claims with confident verdicts,
while voting would discard the conflict.

\paragraph{Relationship to prior frameworks.}
JuryFlow is complementary to panel evaluation~\cite{verga2024poll}: PoLL
aggregates independent votes, whereas JuryFlow adds disagreement-targeted
re-evaluation and a closed correction loop on the same panel. It also differs
from fixed-rubric evaluation~\cite{ye2023flask, kim2024prometheus}. Rather than
using a static rubric for each dimension, JuryFlow induces entries from its own
corrections and updates them over time.

\paragraph{Generality.}
The sequence of targeting, propagation, and rubric induction is not specific to
text quality evaluation. It can apply to multi-agent decision pipelines that
expose disagreement, including data labeling, content moderation, and
multi-agent reasoning.

\section{Limitations and Future Work}
\label{sec:limitations}

\paragraph{The human-guided mode is unvalidated.}
Every result here uses automatic focal selection, so we establish the value of
disagreement-targeted re-evaluation but not that of the human selector
motivating the design. The argument in Section~\ref{sec:discussion} follows from
the construction of the entropy proxy, not from measurement. A user study comparing
human and entropy-based selection in terms of accuracy and effort is therefore
necessary.

\paragraph{Correlated judges.}
Disagreement is informative only when judges err conditionally independently.
Judges sharing training data or exhibiting self-preference
bias~\cite{panickssery2024selfpreference} can reach confident but wrong
consensus, which low entropy will not flag; mitigating such blind spots requires
diversity-aware panel construction.

\paragraph{Cost and rubric drift.}
Cross-instance propagation and a growing rubric raise inference cost and risk
unbounded growth or drift over long runs. Embedding-based deduplication
(Stage~5) mitigates but does not bound these risks. Principled pruning remains
future work.

\paragraph{Scope and evidence.}
Our protocol covers two English benchmarks; generalization to other languages,
modalities, and long-form responses is untested. The reviewer also re-judges
using only the original context. Retrieving external evidence, as in
SAFE~\cite{wei2024safe}, is a natural extension.

\section{Conclusion}
\label{sec:conclusion}

JuryFlow treats inter-judge disagreement as a claim-level signal for allocating
additional evaluation. It scores claims by verdict entropy, re-evaluates the
most contested claims, propagates each correction through a structural graph,
and incorporates the correction into a shared rubric. This process converts a
judge panel into an evaluator that can improve from a single selection. On
MT-Bench and LLMBar, JuryFlow agrees with gold labels more often than
single-judge and majority-vote baselines. Our ablations identify the
contributions of focal re-evaluation, propagation, and rubric induction. Because
entropy ranking replaces human selection in all experiments, future work must
establish what a person contributes beyond this proxy, particularly when the
panel reaches an incorrect consensus.

\begin{acks}
This research was supported by JSPS KAKENHI Grant Number 25K21201.
\end{acks}

\appendix

\section{System Prompt Templates}
\label{app:prompts}

Templates used by each agent; \texttt{\{curly\}} fields are instantiated per
instance.

\paragraph{Judge agent (Stage 1).}
{\footnotesize\ttfamily\raggedright
You are an impartial judge. Decompose the response into atomic, independently
verifiable claims. For each claim, output a verdict (accept / reject /
uncertain), a one-sentence rationale, and the relevant evaluation dimensions
from the fixed tag set \{T\}. Consider the current rubric: \{rubric\}.
Question: \{Q\}. Response: \{A\}. Return JSON matching the schema.\par}

\par\noindent\textbf{Reviewer agent (Stage 4).}
{\footnotesize\ttfamily\raggedright
Re-evaluate the focal claim \{c*\} using only evidence in the question and
response. Weigh supporting and opposing evidence, then output a revised verdict
and an evidence-grounded rationale. Conflicting prior verdicts: \{v\}.\par}

\par\noindent\textbf{Rubric-entry generation (Stage 5).}
{\footnotesize\ttfamily\raggedright
Given the focal claim, its tags, the original conflicting verdicts, and the
revised verdict, write one reusable rubric entry: a criterion description, one
positive example, and one negative example. Output JSON.\par}

\bibliographystyle{ACM-Reference-Format}
\bibliography{references}

\end{document}